\documentclass[final]{cvpr}

\usepackage{times}
\usepackage{epsfig}
\usepackage{graphicx}
\usepackage{amsmath}
\usepackage{amssymb}
\usepackage{multirow}
\usepackage{subfigure}
\usepackage{epstopdf}
\usepackage{booktabs}

\usepackage[pagebackref=true,breaklinks=true,colorlinks,bookmarks=false]{hyperref}

\def\cvprPaperID{11099} % *** Enter the CVPR Paper ID here
\def\confYear{CVPR 2021}

\begin{document}

%%%%%%%%% TITLE
\title{Multi-grained Trajectory Graph Convolutional Networks\\for Habit-unrelated Human Motion Prediction}
\author{Jin Liu$^{1}$, Jianqin Yin$^{1}$\\
$^{1}$Beijing University of Posts and Telecommunications\\%School of Modern Post of Beijing University of Posts and Telecommunications\\
%No.10 Xitucheng Road, Haidian District, Beijing 100876, China\\
{\tt\small \{jinliu, jqyin\}@bupt.edu.cn}
%\and
%Jianqin Yin\\
%Beijing University of Posts and Telecommunications\\%School of Artificial Intelligence\\
%No.10 Xitucheng Road, Haidian District, Beijing 100876, China\\
%{\tt\small jqyin@bupt.edu.cn}
}

\maketitle
%%%%%%%%% ABSTRACT
\begin{abstract}
Human motion prediction is an essential part for human-robot collaboration. Unlike most of the existing methods mainly focusing on improving the effectiveness of spatiotemporal modeling for accurate prediction, we take effectiveness and efficiency into consideration, aiming at the prediction quality, computational efficiency and the lightweight of the model. A multi-grained trajectory graph convolutional networks based and lightweight framework is proposed for habit-unrelated human motion prediction. Specifically, we represent human motion as multi-grained trajectories, including joint trajectory and sub-joint trajectory. Based on the advanced representation, multi-grained trajectory graph convolutional networks are proposed to explore the spatiotemporal dependencies at the multiple granularities. Moreover, considering the right-handedness habit of the vast majority of people, a new motion generation method is proposed to generate the motion with left-handedness,  to better model the motion with less bias to the human habit. Experimental results on challenging datasets, including Human$3.6$M and CMU Mocap, show that the proposed model outperforms state-of-the-art with less than 0.12$\times$ parameters, which demonstrates the effectiveness and efficiency of our proposed method.
\end{abstract}

%%%%%%%%% BODY TEXT
\section{Introduction}

\begin{figure}[t]
\begin{center}
   \includegraphics[width=1\linewidth]{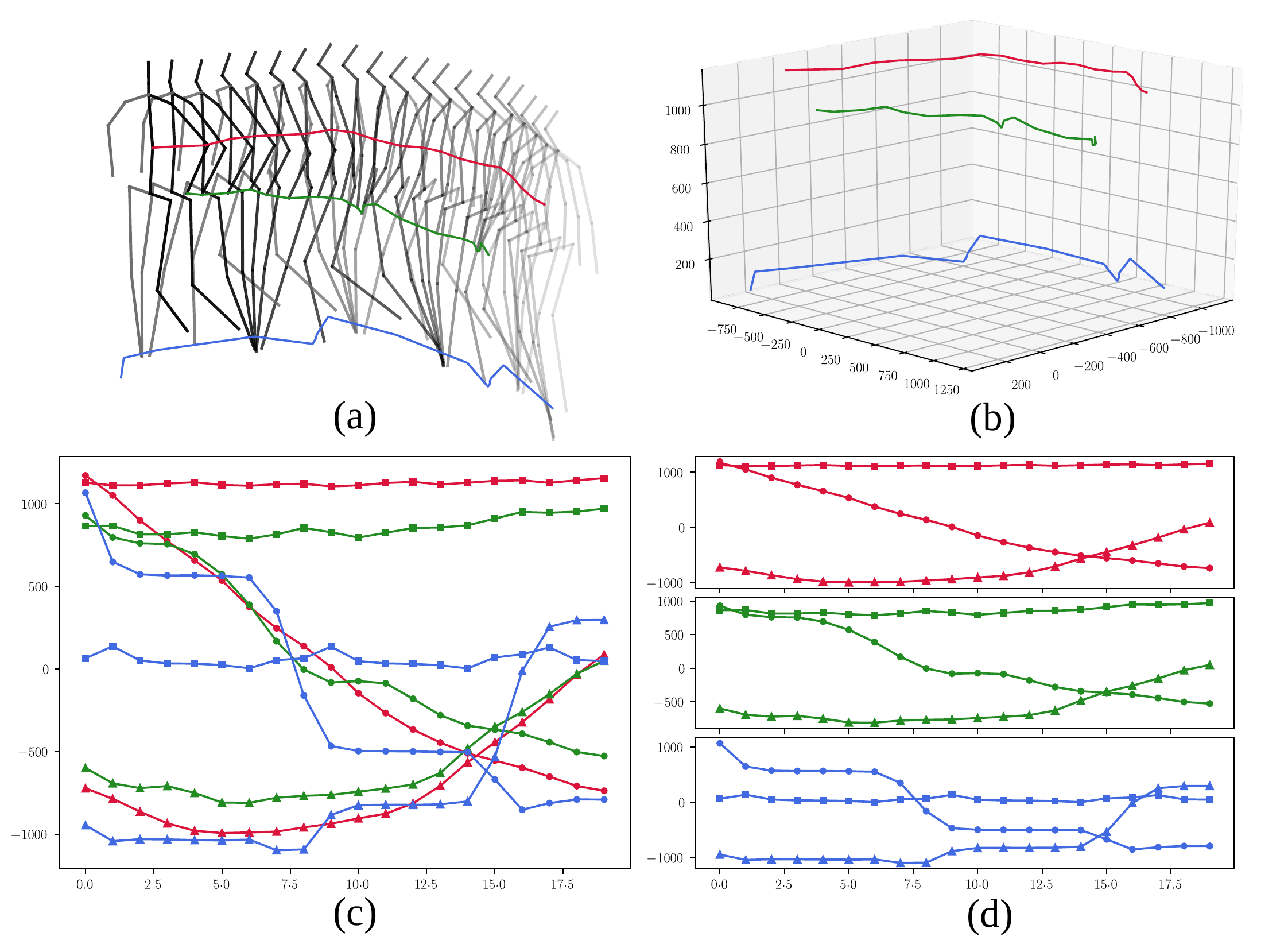}
   \vspace{-2em}
\end{center}
   \caption{The illustration of human motion and multi-grained trajectories. (a) Human motion; (b) Joint trajectories in 3D space; (c) Global sub-joint trajectories; (d) Local sub-joint trajectories.
The curves with the same color belong to the same joint. The curves marked with triangle, point, and square denote the 1D sub-joint trajectories along $x$-axis, $y$-axis, and $z$-axis respectively; the curves without markers are 3D joint trajectories. }
\label{fig:traj}
\vspace{-1em}
\end{figure}

Human motion prediction, aiming at predicting human future poses from observed poses,  has various applications in human-robot collaboration system, such as human-robot interaction \cite{11}, obstacle avoidance \cite{12} and pose tracking \cite{13, a1}. Considering practical scenarios where a robot needs to react to human quickly, human motion prediction algorithms should strike a desired balance between computational efficiency and prediction quality. Human motion can be represented as the evolution of the skeletal sequence. Using the 3D joints based skeletal description, we design a novel motion generation strategy and propose an efficient and lightweight network to achieve the goal.
%The spatial-temporal modeling is the key to generate plausible prediction. Recurrent neural networks (RNNs) based methods \cite{01,02,03} have intrinsic properties to model the temporal information of the time-series signals, but they often suffer from inadequate spatial learning and low computational efficiency \cite{06}. To better mine the spatial dependencies, based on convolutional neural networks (CNNs), \cite{09} put the dimension of joint as channel to capture global spatial co-occurrence features; yet, the global temporal co-occurrence features are hard to mine, due to the convolutional filter only covers partial time snippets. To mine the global temporal co-occurrence features, \cite{07} proposed a new trajectory space and put the dimension of time as channel; however, the global spatial co-occurrence features are hard to mine in this manner. The two CNN-based methods are with a common shortcoming that they cannot capture the global temporal and spatial co-occurrence features simultaneously. This weakness is caused by the fact that CNNs are designed originally to extract features for the data with grid structure and high neighborhood correlation; therefore, for the graph structured human skeleton sequence, the grid shape, limited sizes, and shared weights of convolutional kernels restrain the spatiotemporal co-occurrence feature learning that is crucial for human motion prediction.

The spatial-temporal modeling is the key to generate plausible prediction. Recurrent neural networks (RNNs) based methods \cite{01,02,03} have intrinsic properties to model the temporal information of the time-series signals, but they often suffer from error accumulation \cite{05,07} and low computational efficiency \cite{06,convs2s} due to their recurrent prediction manner. Convolutional neural networks (CNNs) based methods \cite{04,07,05} often capture spatiotemporal features in local-to-global manner, causing their lack of ability to extract global spatiotemporal co-occurrence features. This weakness is caused by the fact that CNNs are designed originally to extract features for the data with grid structure and high neighborhood correlation; therefore, for the graph structured human skeleton, the grid shapes, limited sizes, and shared weights of convolutional kernels restrain the global spatiotemporal co-occurrence feature learning that is crucial for human motion prediction. Currently, graph convolutional networks (GCNs), as a generalization of CNNs, have been wildly applied to skeleton-based human activity analysis \cite{stgcn, 2stream, agc, semantics, 10} and have shown competitive performance. However, most methods modeled spatial, and temporal correlations separately, causing the global spatiotemporal co-occurrence features learning inadequately. Recently, Mao \etal \cite{05} proposed GCNs to explore the spatiotemporal dependencies between trajectories, in which the spatiotemporal modeling is in a coupled manner, and achieved promising performance.
%without the constraints of
%However, the representation of the trajectories is only with one granularity and neglects the inherent relationships between trajectories.
%However, the human motion is represented as single grained trajectories in \cite{05},  which cannot fully express inherent constraints, causing large search space and low efficiency of parameter utilization.
However, the trajectory representation is confined in a single granularity, causing insufficiency of spatiotemporal modeling.

% For the sub-joint trajectories, to the correlations of the trajectories of within one joint is different from the correlations of the different joints.
To solve the aforementioned problems, as shown in Figure \ref{fig:traj}, we represent human motion as multi-grained trajectories. At the joint level, human motion is represented as coarse-grained joint trajectories in 3D space. Then, a joint trajectory graph convolution (JTGC) is proposed to model the spatiotemporal features at the coarse-grained joint trajectory level.
Considering a finer granularity, the spatiotemporal correlations in different directions of motion between joints may be different. For example, for action ``clapping hands'', two hands tend to move in opposite directions horizontally, while moving in the same directions vertically. To explore the fine-grained relationships, we decompose each 3D joint trajectory in the three-dimensional coordinates and propose a fine-grained sub-joint trajectory representation. Specifically, due to the fact that the physiological structures between joints are various, the motion patterns may be quite different for different joints. Therefore, we treat the trajectories within one joint as local sub-joint trajectories and all the sub-joint trajectories as global ones. Then we propose local sub-joint trajectory graph convolution (LSTGC) and global sub-joint trajectory graph convolution (GSTGC) to separately models the local and global spatiotemporal features of the human motion at the sub-joint granularity. Finally, JTGC, LSTGC and GSTGC are fused hierarchically to propose multi-grained trajectory graph convolutional module (MTGCM), which has powerful spatial-temporal modeling ability and can achieve promising performance in a lightweight model manner.

Moveover, due to the habit of using the right hand/leg for the majority of the human being, it is difficult to obtain the motion which are performed using the left hand/leg. Therefore, the data-driven methods tend to learn biases about the habit-related motion patterns, which cannot fit well for the motions executed by the left-handed persons or the motions rarely performed using the left hand/leg by the right-handed persons. In order to augment the adaptation ability to various action execution modes for habit-unrelated motion prediction, we design a mirror transformation scheme inspired by the symmetry of human body. Specifically, due to the fact that human skeleton is on mid-sagittal plane symmetric, we conduct mirror transformation on skeletal data corresponding to the plane. And then we combine the transformed skeletal data and raw skeletal data to train our model. By this simple but effective operation, the prior knowledge of symmetry of human body is introduced and the spatial inherent constraints between left and right limbs are enhanced. Thereby, model training on the augmented data tends to learn more correlations between left and right limbs, and tends to generate habit-unrelated prediction.

%Moveover, due to the habit of using the right hand/leg for the majority of the human being, it is difficult to obtain the motion which are performed using the left hand/leg. In order to augment the adaptation ability to various action execution modes, we design a mirror transformation model at the skeletal level inspired by the symmetry of human body. Specifically, due to the fact that human skeleton is on mid-sagittal plane symmetric, we conduct mirror transformation on skeletal data corresponding to the plane. And then we combine the transformed skeletal data and raw skeletal data to train our model. By this simple but effective operation, the prior knowledge of symmetry of human body is introduced and the spatial inherent constraints between left and right limbs are enhanced. Thereby, model training on the augmented data tends to learn more correlations between left and right limbs.

In conclusion, our contributions are as follows:

%in which inherent relationships between the trajectories are contained explicitly.
1) We represent human motion as multi-grained trajectories, in which coarse-grained joint trajectories and fine-grained sub-joint trajectories are contained. Under this representation, a multi-grained trajectory graph convolutional framework is proposed to effectively and efficiently model spatiotemporal correlations for accurate motion prediction.

%in which prior knowledge of symmetry of human body is introduced
2)	A novel motion generation scheme is designed for habit-unrelated motion prediction; this scheme can effectively generate the motions executed by the left-handed persons or the motions rarely performed using the left hand/leg by the right-handed persons, bring our model learning with less bias to the habit-related motion patterns.

3)	Extensive experiments on Human$3.6$M and CMU Mocap show that the proposed method outperforms the state-of-the-art with much smaller model size, which balances prediction quality computational efficiency.

\section{Related work}

\subsection{Human motion prediction}
%RNNs with the natural property of processing time series, there are numerous methods based on RNNs.
\textbf{RNN-based methods. } Due to the vanilla RNN with the defect that spatial features cannot be well modeled, Fragkiadaki \etal \cite{01} proposed an Encoder-Recurrent-Decoder, in which an encoder and a decoder is introduced before and after their LSTM separately to model spatial dependencies. Jain \etal \cite{02} proposed a Structural-RNN to further model high-level spatio-temporal information for prediction. Yet, the two methods suffer from the discontinuity between predicted frames \cite{03}. A framework in which GRUs with residual connection was proposed by Martinez \etal \cite{03} to alleviate the discontinuous problem. Another shortcoming of RNNs is that they tend to forget the dependencies in long term. Tang \etal \cite{14} introduced a temporal attention before their recurrent module to enhance long-term temporal modeling. Although promising predictions they made, the defects of low computational efficiency become obstacles for RNN-based algorithms to practical application.

\textbf{Feedforward-based methods. } To overcome the problem of low efficiency and lack of spatial modeling capability of RNN-based methods, feedforward-based methods, like fully-connected-based \cite{08}, CNN-based \cite{04,07} and GCN-based \cite{05,10}, are proposed successively. Bütepage \etal \cite{08} proposed an autoencoder to model temporal information and hierarchical fully-connected networks are proposed to capture spatial dependencies. Li \etal \cite{04} proposed convolutional networks to human motion prediction in which different sizes of the filters are used to capture multi-scale spatial relations. The spatial modeling of the method depends heavily on the size of filters and cannot capture global spatial features in one layer. Liu \etal \cite{07} proposed using CNNs map the pose to trajectory space, and model the spatial-temporal dependencies in the trajectory space. Although global temporal co-occurrence features were captured in the trajectory space, the spatial modeling also relies on the expending of receptive field of CNNs and also cannot capture global spatial features in one layer. Mao \etal \cite{05} adopted the Discrete Cosine Transform (DCT) to map trajectories from pose space to frequency domain to model temporal correlations, and GCNs were employed to model the spatial dependencies and achieved state-of-the-art performance. However, the temporal modeling based on the manual features is not flexible enough \cite{07} and may cause the insufficiency of the corresponding spatial modeling. Li \etal \cite{10} proposed GCN-based encoder with multiscale graphs to extract spatial-temporal features and a graph-based GRU decoder was used to generate prediction with recurrent manner. Although the efficient spatiotemporal dependencies the decoder modeled, the RNN-based decoder also suffers from low computational efficiency.

\subsection{Human motion generation}

In the field of human motion prediction, few works \cite{01,02,15} considered the generation of the skeletal data. Fragkiadaki \etal \cite{01} proposed to add zero mean Gaussian noise to the skeletal data to improve the generalization of their model. And the variance of the noise is gradually increased during training. This strategy is also used in \cite{02,15}. However, Martinez \etal \cite{03} pointed out that the noise added to the input is difficult to tune and degrades the quality of the prediction. In addition, with the increase of the noise variance, the structure of skeleton is corrupted, such as the change of bone length. In order to model spatial constraints of skeleton, Ghosh \etal \cite{15} utilized dropout with increased drop probability to discard limbs in skeleton randomly, and then a dropout autoencoder was proposed to recover the original data. Yet, the method requires redundant training process of the autoencoder and the training of the autoencoder and predictor is not in end-to-end manner. In addition, the way of discarding limbs randomly destroys the integrity of the skeleton. It is described in \cite{15} that the recovered data are not identical with the original data for some complex poses. Therefore, the predicted poses based on the biased data recovered by the autoencoder may deviate from the real poses.

In summary, the above methods of skeletal data augmentation not consider the nature of inherent structure of human body. On the other hand, for the skeletal data, there are few generic data augmentation methods like image augmentation. More importantly, due to the habit of using the right hand/leg for the majority of the human being, it is difficult to obtain the motion which are performed using the left hand/leg. These motivate us to take the structural properties of human body into consideration and design a generic method for the augmentation of the skeletal data to enhance spatial inherent constraints and expand the data with small sample that is hard to collect in the practical scene.

\section{Method}

\begin{figure*} %framework
\begin{center}
   \includegraphics[width=01\linewidth]{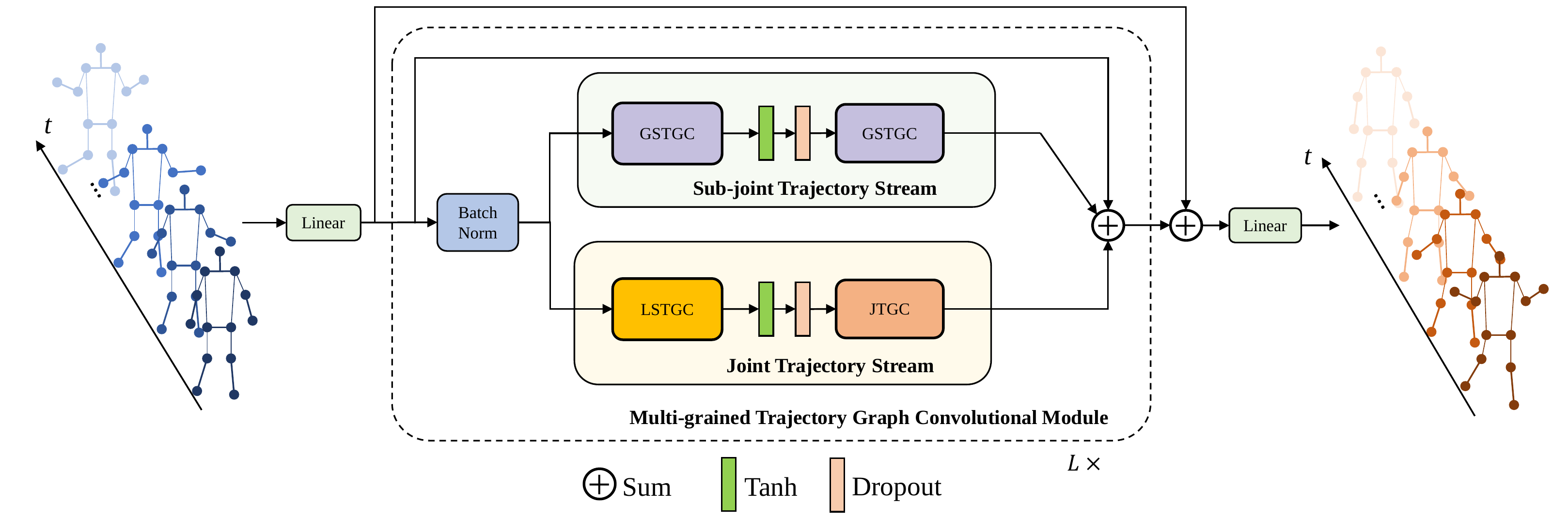}
\end{center}
   \caption{The architecture of our proposed model.  Firstly, a linear layer is employed to transform each sub-joint trajectory to a new semantic space. Then, we stack the Multi-grained Trajectory Graph Convolutional Module (MTGCM) with $L$ layers to mine spatiotemporal features in the semantic space. Finally, a linear layer is used to generate prediction by mapping the spatiotemporal features from the semantic space to the pose space. }%The architecture of our proposed model. Firstly, a linear layer is used to transform each joint projection trajectory to a new semantic space. Then, we stack the XXXXX with $L$ layers to mine spatiotemporal dependencies in the semantic space. Finally, a linear layer is employed to generate prediction by mapping the spatiotemporal dependencies from the semantic space to the pose space%.}
\label{fig:framework}
\end{figure*}

\subsection{Problem formulation}

Given an observed motion sequence  ${\bf{F = }}\left\{ {{{\bf{f}}_1}{\bf{,}}{{\bf{f}}_2}{\bf{,}}...{\bf{,}}{{\bf{f}}_T}} \right\}$  with $T$  frames of skeletal data, the goal of human motion prediction is to predict the future motion sequence with $T'$  frames generating ${\bf{F'}} = \left\{ {{{\bf{f}}_{T + 1}},{{\bf{f}}_{T + 2}},...,{{\bf{f}}_{T + T'}}} \right\}$, where ${{\bf{f}}_i} \in {{\mathbb{R}}^{J \times D}}$ , $J$ is the number of joints in human skeleton, $D=3$ denotes 3D position of a joint. Naturally, the human motion is the evolution of the graph structured human poses over time and thus both intra-frame and inter-frame modeling, \ie spatial-temporal modeling, are essential to human motion prediction.

\subsection{Representation of human motion}

Human motion can be regarded as a set of trajectories and the representation of the trajectories has multiple granularities. Below we introduce a series of multi-grained trajectories to represent human motion; then corresponding graphs are proposed following the multi-grained trajectories to model their dependencies.
%Human motion is a complex system with multiple granularities. The representation of human skeleton has multiple granularities. In general, a skeleton is presented by joints, and the human motion can be regarded the evolution of the group of the trajectories of 3D joints, which is a representation at the joint level; Furthermore, each joint is represented as a position in the 3D space with three components, i.e. x, y, z, which is a representation at the sub-joint granularity. From the perspective of sub-joint, the human motion is the evolution of the group of the trajectories of 1D sub-joints. That is to say, the human motion is a multi-level system with joint and sub-joint granularity. To adequately mine spatial-temporal correlations at the multiple granularities, we propose three types of GCNs.
%The human motion sequence can be described from different perspectives, and in different perspectives, there are different granularities of spatiotemporal correlations can be explored. Next, from two perspectives, we introduce two trajectories with different granularities to represent human motion.

\textbf{Joint trajectory and its graph. }%\subsubsection{Joint trajectory and its graph}%\textbf{Joint trajectory and its graph. }
Generally, a skeleton ${\bf{f}}_i$  is presented by a set of joints in 3d space, and human motion $\bf{F}$ can be regarded as the evolution of a group of trajectories of the joints, \ie $\left\{ {{{\bf{J}}_1},{{\bf{J}}_2},...,{{\bf{J}}_J}} \right\}$,  where ${{\bf{J}}_i} \in {\mathbb{R}^{3 \times T}}$ represents the trajectory of $i$th joint in 3D space. Then, based on the joint trajectories, we construct a type of joint trajectory graph $\mathcal{G}^{\mathrm{JT}}=\left(\mathcal{V}^{\mathrm{JT}}, \mathcal{E}^{\mathrm{JT}}\right)$ in which the set of vertex $\mathcal{V}^{\mathrm{JT}}=\left\{\mathbf{J}_{1}, \mathbf{J}_{2}, \ldots, \mathbf{J}_{J}\right\}$ includes all the joint trajectories, and the set of edge $\mathcal{E}^{\mathrm{JT}}=\left\{C_{i j}^{\mathrm{JT}} \mid i, j=1,2, \ldots, J\right\}$ contains pair-wise connectivity of the joint trajectories.  Under the representation, the spatiotemporal correlations between joint trajectories can be expressed.

%Note that we treat human motion as a graph.
%A joint is a physical entity that can be described as a position in 3D space.
%A joint trajectory can be decomposed into three trajectories along $x$, $y$, $z$ directions.

%\subsubsection{Global sub-joint trajectory and its graph}%\textbf{Global sub-joint trajectory and its graph. }
%The correlations in different directions of motion is distinct, which is with finer granularity compared with the above joint trajectories. For example, for action ``clapping hands'', two hands tend to move in opposite directions horizontally, while moving in the same directions vertically. Based on the observation, from the perspectives of directions of motion, we proposed sub-joint trajectories by decomposing the joint trajectories in different directions of motion.
\textbf{Global sub-joint trajectory and its graph. }The spatiotemporal correlations in different directions of motion between joints may be different. Inspired by \cite{05},  we decompose all the 3D joint trajectories in the three-dimensional coordinates and propose global sub-joint trajectories to represent human motion. In detail, we decompose each joint trajectory ${\bf{J}}_i$ in three directions (\ie $x$, $y$, $z$), generating sub-joint trajectories $\left\{\mathbf{J}_{i}^{x}, \mathbf{J}_{i}^{y}, \mathbf{J}_{i}^{z}\right\}$ , where $\mathbf{J}_{i}^{x}, \mathbf{J}_{i}^{y}, \mathbf{J}_{i}^{z} \in {{\mathbb{R}}^{T}}$ represents the trajectory of the $i$th joint in $x$, $y$, $z$ direction respectively. Further, the human motion $\bf{F}$ can be denoted as the evolution of a set of the sub-joint trajectories  $\left\{\mathbf{J}_{1}^{x}, \mathbf{J}_{1}^{y}, \mathbf{J}_{1}^{z}, \ldots, \mathbf{J}_{J}^{x}, \mathbf{J}_{J}^{y}, \mathbf{J}_{J}^{z}\right\}$, called global sub-joint trajectories.

% In order to distinguish the sub-joint trajectories within one joint or not,  we call the joint trajectory within one joint and the joint trajectory of all joints
% Based on the sub-joint trajectory, we propose two graphs.
%\textbf{Global sub-joint trajectory graph. }
To mine the spatiotemporal correlations between the global sub-joint trajectories, we construct a global sub-joint trajectory graph $\mathcal{G}^{\mathrm{GS}}=\left(\mathcal{V}^{\mathrm{GS}}, \mathcal{E}^{\mathrm{GS}}\right)$, in which the set of vertex $\mathcal{V}^{\mathrm{GS}}=\left\{\mathbf{J}_{1}^{x}, \mathbf{J}_{1}^{y}, \mathbf{J}_{1}^{z}, \ldots, \mathbf{J}_{J}^{x}, \mathbf{J}_{J}^{y}, \mathbf{J}_{J}^{z}\right\}$ includes all the sub-joint trajectories, and the set of edge $\mathcal{E}^{\mathrm{GS}}=\left\{C_{i j}^{\mathrm{GS}} \mid i, j=1,2, \ldots, N\right\}$ contains pair-wise connectivity of the vertexes, where $N=J \times 3$ denotes the number of all the sub-joint trajectories. Compared with the above joint trajectory graph $\mathcal{G}^{\mathrm{JT}}$, the global sub-joint trajectory graph $\mathcal{G}^{\mathrm{GS}}$ is with more flexible ability to model the spatiotemporal dependencies at a finer granularity.

\textbf{Local sub-joint trajectory and its graph. }
%Specifically, we decompose each joint trajectory ${\bf{J}}_i$ in three directions (\ie $x$, $y$, $z$), generating sub-joint trajectories $\left\{\mathbf{J}_{i}^{x}, \mathbf{J}_{i}^{y}, \mathbf{J}_{i}^{z}\right\}$ , where $\mathbf{J}_{i}^{x}, \mathbf{J}_{i}^{y}, \mathbf{J}_{i}^{z} \in {{\mathbb{R}}^{T}}$ represents the trajectory of the $i$th joint in $x$, $y$, $z$ direction respectively.
The global sub-joint trajectory graph treats every sub-joint trajectory indiscriminately, either within one joint or not. However, due to the fact that the physiological structures between joints are various, the motion pattern may be quite different between joints, such as bending direction and extent of joints. For example, in action ``squatting down'', the bending directions of knees and hips are completely opposite. To discriminate each joint and distinguish the motion patterns between joints, we treat the sub-joint within one joint, \ie $\left\{\mathbf{J}_{i}^{x}, \mathbf{J}_{i}^{y}, \mathbf{J}_{i}^{z}\right\}$, as local sub-joint trajectories. And based on the local sub-trajectory representation, we construct a local sub-joint graph $\mathcal{G}_{i}^{\mathrm{LS}}=\left(\mathcal{V}_{i}^{\mathrm{LS}}, \mathcal{E}_{i}^{\mathrm{LS}}\right)$ for each joint trajectory ${\bf{J}}_i$ separately, where the set of vertex $\mathcal{V}_{i}^{\mathrm{LS}}=\left\{\mathbf{J}_{i}^{x}, \mathbf{J}_{i}^{y}, \mathbf{J}_{i}^{z}\right\}$ includes three sub-joint trajectories within one joint, and the set of edge $\mathcal{E}_{i}^{\mathrm{SL}}=\left\{C_{i j}^{\mathrm{SL}} \mid i, j=1,2,3\right\}$ contains pair-wise connectivity of the vertexes. Then a set of local sub-joint trajectory graph can be defined as $\left\{\mathcal{G}_{1}^{\mathrm{SL}}, \mathcal{G}_{2}^{\mathrm{SL}}, \ldots, \mathcal{G}_{J}^{\mathrm{SL}}\right\}$ to discriminate each joint.

\begin{figure}[t] %module
\begin{center}
   \includegraphics[width=1\linewidth]{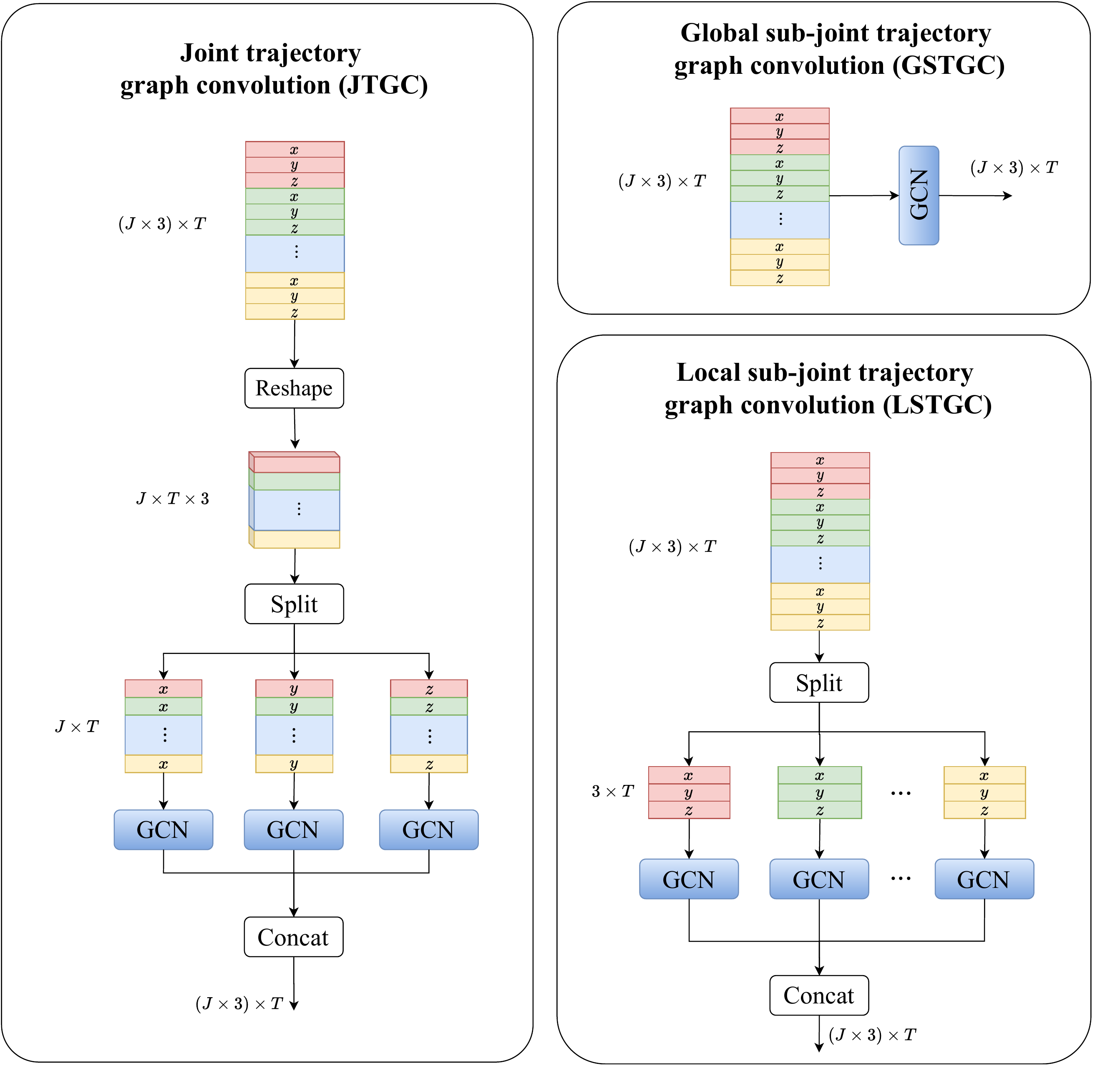}
\end{center}
   \caption{The implementation of Joint Trajectory Graph Convolution (JTGC), Global Sub-joint Trajectory Graph Convolution (GSTGC) and Local Sub-joint Trajectory Graph Convolution (LSTGC). }
\label{fig:module}
\vspace{-1.0em}
\end{figure}

\subsection{Graph convolution based on multi-grained trajectories}

To extract multi-grained spatiotemporal features, we propose three types of graph convolutions based on the above three kinds of graphs separately, joint trajectory graph, global sub-joint trajectory graph and local sub-joint trajectory graph. The illustration of the modules is shown in Figure \ref{fig:module}.%\ie joint trajectory graph $\mathcal{G}^{\mathrm{JT}}$, global sub-joint trajectory graph $\mathcal{G}^{\mathrm{GS}}$, and a group of local sub-joint trajectory graphs $\mathcal{G}_{i}^{\mathrm{LS}}  \left(i=1,2, \ldots, J\right)$

\textbf{Joint trajectory graph convolution (JTGC). }
For the joint trajectory graph $\mathcal{G}^{\mathrm{JT}}$, with vertex set $\mathcal{V}^{\mathrm{JT}}=\left\{\mathbf{J}_{1}, \mathbf{J}_{2}, \ldots, \mathbf{J}_{J}\right\}$ and edge set $\mathcal{E}^{\mathrm{JT}}=\left\{C_{i j}^{\mathrm{JT}} \mid i, j=1,2, \ldots, J\right\}$, the connective strength between the node ${\bf{J}}_{i}$ and ${\bf{J}}_{j}$ is ${C_{i j}^{\mathrm{JT}}}$. Then, the pair-wise connective strength between all vertexes can be defined as a weighted adjacent matrix $\mathbf{A}^{\mathrm{JT}} \in \mathbb{R}^{J \times J}$ with element ${C_{i j}^{\mathrm{JT}}}$ in $i$th row $j$th column of the matrix. Inspired by \cite{05}, to adaptively mine the potential dependencies between vertexes, the adjacent matrix is learnable during optimization. The vertex set is reformulated as a tensor $\mathbf{V}^{\mathrm{JT}}=\left[\mathbf{J}_{1}, \mathbf{J}_{2}, \ldots, \mathbf{J}_{J}\right] \in \mathbb{R}^{J \times 3 \times T}$. Then, the graph convolution based on the joint trajectory graph can be formulated as:
\begin{equation}
\mathbf{Y}^{\mathrm{JT}}=\|_{i=1}^{3} \mathbf{A}^{\mathrm{JT}} \mathbf{V}_{:, i,:}^{\mathrm{JT}} \mathbf{W}^{\mathrm{JT}}
\end{equation}
where ``$||$'' denotes the operation of tensor concatenation, $\mathbf{V}_{:, i,:}^{\mathrm{JT}} \in \mathbb{R}^{J \times T}$, %$\mathbf{W}^{\mathrm{JT}} \in \mathbb{R}_{T \times 3 \times H}$ is a learnable weight,
$\mathbf{Y}^{\mathrm{JT}} \in \mathbb{R}^{\left(J \times 3\right) \times T}$ is the output of the graph convolution.

\textbf{Global sub-joint trajectory graph convolution (GSTGC). }
The spatiotemporal correlations in different directions of motion are distinct. To model the fine-grained spatiotemporal correlations with more flexibility, a graph convolution based on sub-joint trajectory is proposed. For the global sub-joint trajectory graph $\mathcal{G}^{\mathrm{GS}}$, similar to the STGC, global sub-joint trajectory graph can be defined as:
\begin{equation}
\mathbf{Y}^{\mathrm{GS}}=\mathbf{A}^{\mathrm{GS}}\mathbf{V}^{\mathrm{GS}} \mathbf{W}^{\mathrm{GS}}
\end{equation}
where $\mathbf{V}^{\mathrm{GS}} \in \mathbb{R}^{\left(J \times 3\right)\times T}$, and $\mathbf{A}^{\mathrm{GS}} \in \mathbb{R}^{N\times N}$ represents the adjacency matrix of the global sub-joint trajectory graph, and each element is learnable and reflects the pair-wise connective strength between two sub-joint trajectories.

\textbf{Local sub-joint trajectory graph convolution (LSTGC). }
To discriminate each joint and distinguish the motion patterns between joints, for each local sub-joint trajectory graph $\mathcal{G}_{i}^{\mathrm{LS}}$, we define a sub-joint trajectory graph convolution:
\begin{equation}
\mathbf{Y}^{\mathrm{LS}}=\|_{i=1}^{J}\mathbf{A}_{i}^{\mathrm{LS}}\mathbf{V}_{i}^{\mathrm{LS}} \mathbf{W}^{\mathrm{LS}}
\end{equation}
where $\mathbf{V}_{i}^{\mathrm{LS}} \in \mathbb{R}^{3\times T}$ , and $\mathbf{A}_{i}^{\mathrm{LS}} \in \mathbb{R}^{3\times 3}$ denotes the adjacency matrix of the local sub-joint trajectory graph, and each element is learnable and reflects the pair-wise connective strength between two sub-joint trajectories within $i$th joint.

%In summary, the three types of graph convolutions are all with their merits. In detail, the GSTGC explore the spatiotemporal correlations between the global sub-joint trajectories, which is beyond constraints of joints and can flexibly mines the spatiotemporal correlations at a finer granularity. The LSTGC can model the spatiotemporal features between the local sub-joint trajectories, which is at intra-joint level and the spatiotemporal modeling is constrained to one joint. For the JTGC, the coarse-grained spatiotemporal dependencies between the joint trajectories can be captured, which is at inter-joint level.

\subsection{Multi-grained Trajectory Graph Convolutional Module}

%Based on the above three types of graph convolution, we propose multi-grained trajectory graph convolutional module (MTGCM), which contains two streams, as shown in Figure \ref{fig:framework}. Specifically, one stream consists of two GSTGC blocks; due to the GSTGC is with the ability of exploring the spatiotemporal correlations beyond constraints of joints, we call the stream as spatial non-constrained stream (SNCS). In other stream, the LSTGC and the JTGC are introduced  to mine the spatiotemporal correlations from intra-joint to inter-joint, which is under the constraint of joint construction, and thus we call the stream as spatial constrained stream (SCS).The spatiotemporal modeling with no constraints and under constraints are both crucial for human motion prediction; therefore, we fuse the SNCS and SCS to fully mine the spatiotemporal correlations for accurate prediction.
%due to the GSTGC is with the ability of exploring the spatiotemporal correlations beyond constraints of joints,
%For the SCS, we employ the LSTGC and the JTGC to mine the spatiotemporal from intra-joint to inter-joint  spatial constraints.
To mine multi-grained spatiotemporal features, based on the multi-grained trajectory representation, we propose a multi-grained trajectory graph convolutional module (MTGCM) by using the above three types of graph convolutions. As shown in Figure \ref{fig:framework}, a batch normalization (BN) layer is introduced to normalize data for stable training. Then, we employ two streams, including sub-joint trajectory stream and joint trajectory stream, to model spatiotemporal correlations at different granularities. Moreover, a residual connection \cite{resnet} is added in the MTGCM.

%the Based on the above three types of graph convolutions, we proposed a multi-grained trajectory graph convolutional module (MTGCM), which contains two streams, as shown in Figure \ref{fig:framework}.
%based on the sub-joint trajectory representation of human motion, two GSTGC blocks are employed to capture fine-grained spatiotemporal dependencies in the SJTS stream.
%The correlations in different directions of motion is distinct and the GSTGC is with the ability of capturing the fine-grained correlations; and a nonlinear activation and a dropout layer are used between the two blocks. The SJTS aims to extract fine-grained spatiotemporal features.

\textbf{Joint Trajectory Stream (JTS). }
To capture the spatiotemporal correlations between joint trajectories, we propose a JTS in which a LSTGC and JTGC layer are stacked. In detail, the LSTGC is to capture the capture the local spatiotemporal features within one joint and the JTGC is after it to further mine global spatiotemporal features.

\textbf{Sub-joint Trajectory Stream (SJTS). }
Due to the correlations in different directions of motion is distinct, based on the sub-joint trajectory representation of human motion, we built the SJTS in which two GSTGC blocks are employed to capture the fine-grained spatiotemporal dependencies.

%where $\mathbf{A}^{\mathrm{LS}} \in \mathbb{R}^{3\times 3}$   represents the adjacency matrix of the global sub-joint trajectory graph, and each element is learnable and reflects the pair-wise connective strength between two sub-joint trajectories within one joint; and $\mathbf{V}^{\mathrm{LS}} \in \mathbb{R}^{3\times T}$ with 3 the number of the sub-joint trajectories within one joint.
\subsection{Network architecture}
Based on the MTGCM, we propose an end-to-end multi-grained trajectory graph convolutional networks for human motion prediction. The pipeline of our model is shown in Figure \ref{fig:framework}.
The model consists of three parts. In detail, inspired by the \cite{07}, an input layer with a linear transformation is introduced to map each global sub-joint trajectory from pose space to a trajectory space with the temporal semantic information encoded in the process, \ie ${\bf{E = F}}{{\bf{W}}_\mathrm{in}}$, where ${\bf{F}} \in {{\mathbb{R}}^{N \times T}}$ is the observed motion sequence with $N$ the number of global sub-joint trajectories, and ${{\bf{W}}_\mathrm{in}} \in {{\mathbb{R}}^{T \times H}}$ is a learnable parameter with $H$ the number of hidden neurons; in the middle layer, the MTGCM is stacked by $L$ layers to extract multi-grained spatiotemporal features in the semantic space;  at the end of the model, an output layer with a linear transformation is employed to map the spatial-temporal features to the pose space to yield predictions.% \ie ${\bf{F'}} = {\bf{E}}{{\bf{W}}_\mathrm{out}} + {\bf{b}}_\mathrm{out}$. %Here, there is a trivial detail that the linear transformation in input layer not adds bias only due to the batch normalization is used after that.
%\subsection{Network architecture}
%
%Based on the above three types of graph convolution, we propose an end-to-end model with light parameters to predict human motion. The network consists of three parts: (1) the input layer with a linear transformation to map each sub-joint trajectory on skeletal sequence from pose space to a trajectory space with a temporal semantic information encoded in the space, \ie ${\bf{E = F}}{{\bf{W}}_{in}}$, where ${\bf{F}} \in {{\mathbb{R}}^{N \times T}}$ and ${{\bf{W}}_{in}} \in {{\mathbb{R}}^{T \times D}}$ is a learnable parameter; (2) the global spatiotemporal coupled graph convolutional module stacked by $L$ layers in the middle layer to mine coupled spatial-temporal dependencies in the semantic space; (3) the output layer with a linear transformation to map the spatial-temporal dependency to the pose space to yield prediction, \ie ${\bf{F'}} = {\bf{E}}{{\bf{W}}_{out}} + {\bf{b}}_{out}$. Here, there is a trivial detail that the linear transformation in input layer not adds bias only due to the batch normalization is used after that. Nothing special for input and out layer, and next, we will focus on the global spatiotemporal coupled graph convolutional module.

\subsection{Mirror transformation for habit-unrelated prediction}

Considering the context of the habitation of using the right hand/leg for the majority of the human being, it is difficult to obtain the motion which are performed using the left hand/leg. To generate the data with small sample that is hard to collect in the practical scene, we design a novel skeletal data augmentation method based on mirror transformation of the skeletal data inspired by the symmetry of human body, which is important to make the data-driven model learning with less bias and improve the generalization ability for new subjects with different motion habits.

\begin{figure}[t]
\begin{center}
   \includegraphics[width=1\linewidth]{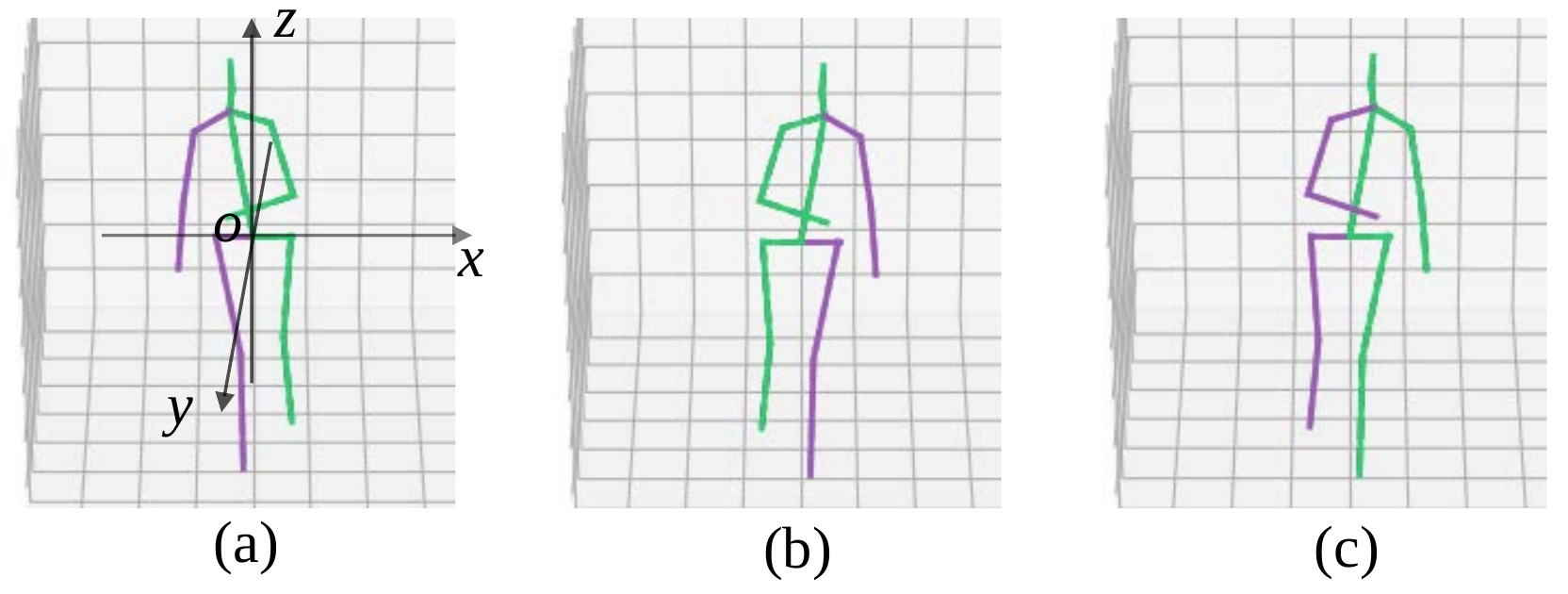}
\end{center}
   \caption{ Mirror transformation. (a) is the visualization of raw skeletal data; (b) represents the coordinate mirror transformation of (a) corresponding to the plane $yOz$; (c) represents the structure mirror transformation of (b).}
\label{fig:mirror}
\end{figure}

We remove the global translations and rotations and put the origin of the coordinate system on the hip of skeleton, with the human body always faces the $y$-axis and the vertical direction is the $z$-axis. After the preprocessing, as shown in Figure \ref{fig:mirror}, the human skeleton is on  the plane $yOz$  symmetry. For a 3D skeleton ${\mathbf{f}} \in \mathbb{R}^{J \times 3}$, first, we take mirror transformation of ${\mathbf{f}}$  corresponding to the plane  $yOz$ generating $\mathbf{g}$ :
\begin{equation}
 {\mathbf{g}} = \mathbf{f}
\left[ {\begin{array}{*{20}{c}}
{\begin{array}{*{20}{c}}
{ - 1}\\
0\\
0
\end{array}}&{\begin{array}{*{20}{c}}
0\\
1\\
0
\end{array}}&{\begin{array}{*{20}{c}}
0\\
0\\
1
\end{array}}
\end{array}} \right]
\end{equation}

Then we exchange left limbs and right limbs of the mirror transformed skeletal data $\mathbf{g}$  getting the augmented skeletal data ${\mathbf{f}}_{mt}$ :
 \begin{equation}
  {\mathbf{f}}_{mt} = \left( {\mathop \Pi \limits_{i = 1}^M {{\mathbf{I}}_{{l_i}{r_i}}}} \right)\mathbf{g}
\end{equation}
where ${l_i}, {r_i} (i=1,2,\ldots, M)$  denote the indexes of the symmetrical joints in left and limbs of matrix $\mathbf{g}$, respectively, \eg left elbow and right elbow, and $M$ denotes the paired number of the symmetrical joints, and ${\mathbf{I}}_{mn}$ represents an elementary matrix acquired by the exchange of the $m$th row and the $n$th row of the identity matrix.

%------------------------------------------------------------------------
\section{Experiments}
\begin{table*}[!t] % short-term results on h36m
\caption{Short-term prediction on H$3.6$M. Where ``ms'' denotes ``milliseconds''.}
\scriptsize
\begin{center}
\begin{tabular}{c|cccc|cccc|cccc|cccc}%{p{.9cm}p{0.18cm}p{0.18cm}p{0.3cm}p{0.3cm}p{0.3cm}p{0.35cm}|p{0.18cm}p{0.18cm}p{0.3cm}p{0.3cm}p{0.3cm}p{0.35cm}|p{0.18cm}p{0.18cm}p{0.3cm}p{0.3cm}p{0.3cm}p{0.35cm}|p{0.18cm}p{0.18cm}p{0.3cm}p{0.3cm}p{0.3cm}p{0.35cm}}
\hline
motion & \multicolumn{4}{c}{Walking} & \multicolumn{4}{c}{Eating}& \multicolumn{4}{c}{Smoking} & \multicolumn{4}{c}{Discussion}\\
\hline
time(ms)&80&160&320&400&80&160&320&400&80&160&320&400&80&160&320&400 \\
\hline
ResSup\cite{03} &23.8 &40.4& 62.9& 70.9& 17.6& 34.7& 71.9& 87.7& 19.7& 36.6& 61.8& 73.9&31.7& 61.3& 96.0& 103.5 \\
ConvS2S\cite{04} &17.1 &31.2&53.8&61.5&13.7&25.9&52.5&63.3&11.1&21.0&33.4&38.3&18.9&39.3&67.7&75.7\\
%\hline
 LTD\cite{05}&8.9 &15.7&{\bf 29.2}&{\bf 33.4}& 8.8& 18.9& 39.4& {\bf 47.2}& 7.8& 14.9& 25.3&{28.7}& 9.8& 22.1&{39.6} &{\bf 44.1} \\
\hline
 Ours&{\bf 7.9} & {\bf 14.8} &29.5& 34.8& {\bf 7.9} &{\bf 18.0}&{\bf 38.9}&{47.7} &{\bf 6.1}&{\bf 12.6} &{\bf 23.3}&{\bf 28.4}&{\bf 7.7} &{\bf 18.9}&{\bf 38.2}&{44.9}\\
\hline
\hline
motion & \multicolumn{4}{c}{Directions} & \multicolumn{4}{c}{Greeting}& \multicolumn{4}{c}{Phoning} & \multicolumn{4}{c}{Posing}\\
\hline
time(ms)&80&160&320&400&80&160&320&400&80&160&320&400&80&160&320&400 \\
\hline
ResSup\cite{03} & 36.5 &56.4& 81.5& 97.3&37.9& 74.1& 139.0& 158.8 &25.6& 44.4& 74.0& 84.2& 27.9& 54.7& 131.3& 160.8 \\
ConvS2S\cite{04} & 22.0&37.2 &59.6& 73.4 &24.5 &46.2 &90.0& 103.1& 17.2& 29.7& 53.4 &61.3& 16.1& 35.6& 86.2& 105.6\\
%\hline
LTD\cite{05}& 12.6 & 24.4&{48.2}&{ 58.4}& 14.5& 30.5& 74.2& 89.0 & 11.5& 20.2& 37.9& {\bf 43.2}&9.4& 23.9& {66.2}&{ 82.9}\\
\hline
Ours&{\bf 9.5}&{\bf 20.9}&{\bf 46.8}&{\bf 57.5}&{\bf 13.0}&{\bf 27.2}&{\bf 64.1}&{\bf 80.9}& {\bf 11.1}&{\bf 19.1}& {\bf 37.5}&{43.9}  &{\bf 7.1}&{\bf 21.0}&{\bf 63.2}&{\bf 79.2}\\
\hline
\hline
motion & \multicolumn{4}{c}{Purchases} & \multicolumn{4}{c}{Sitting}& \multicolumn{4}{c}{Sitting Down} & \multicolumn{4}{c}{Taking Photo}\\
\hline
time(ms)&80&160&320&400&80&160&320&400&80&160&320&400&80&160&320&400 \\
\hline
ResSup\cite{03} & 40.8& 71.8& 104.2& 109.8 &34.5& 69.9& 126.3 &141.6& 28.6& 55.3& 101.6& 118.9 &23.6 &47.4& 94.0& 112.7\\
ConvS2S\cite{04} & 29.4& 54.9& 82.2& 93.0 &19.8 &42.4& 77.0& 88.4& 17.1& 34.9& 66.3& 77.7& 14.0& 27.2& 53.8& 66.2\\
%\hline
LTD\cite{05}& 19.6&38.5& 64.4& 72.2&10.7& 24.6& 50.6&62.0&11.4 &27.6& 56.4& 67.6& 6.8& 15.2& {\bf 38.2}&{\bf 49.6} \\
\hline
Ours&{\bf 17.2}&{\bf 37.0}&{\bf 58.1}&68.2& {\bf 8.9}&{\bf 21.5}&{\bf 44.6}&{\bf 55.9}& {\bf 10.8}&{\bf 26.6}&{\bf 49.2}&{\bf 59.6}&{\bf  5.5}&{\bf 14.5}&38.5&50.2\\
\hline
%\hline
%motion & \multicolumn{4}{c}{Walking} & \multicolumn{4}{c}{Eating}& \multicolumn{4}{c}{Smoking} & \multicolumn{4}{c}{Average}\\
%\hline
%time(ms)&80&160&320&400&80&160&320&400&80&160&320&400&80&160&320&400 \\
%\hline
%ResSup\cite{03} & 29.5& 60.5& 119.9& 140.6& 60.5& 101.9& 160.8& 188.3& 23.5& 45.0& 71.3& 82.8& 30.8& 57.0& 99.8& 115.5\\
%ConvS2S\cite{04} &17.9& 36.5& 74.9& 90.7& 40.6& 74.7& 116.6& 138.7& 15.0& 29.9& 54.3& 65.8& 19.6& 37.8& 68.1& 80.2\\
%%\hline
%DTraj\cite{05}& 9.5& 22.0& 57.5& 73.9& 32.2& 58.0& 102.2& 122.7 & 8.9& { 18.4}& 35.3& 44.3& 12.1& 25.0& 51.0& 61.3\\
%\hline
%Ours&{\bf 7.6}&{\bf 19.7}&{\bf 53.6}&{\bf 69.3} & {\bf 23.5}&{\bf 51.9}&{\bf 98.4}&{\bf 116.7}&  {\bf 7.5}&{\bf 16.5}&{\bf 32.2}&{\bf 42.2}&{\bf 10.1}&{\bf 22.7}&{\bf 47.7}&{\bf 58.6}\\
%\hline
\end{tabular}
\begin{tabular}{p{1.1cm}|p{0.42cm}p{0.42cm}p{0.42cm}p{0.42cm}|p{0.42cm}p{0.42cm}p{0.42cm}p{0.42cm}|p{0.42cm}p{0.42cm}p{0.42cm}p{0.42cm}|p{0.42cm}p{0.42cm}p{0.42cm}p{0.42cm}|p{1cm}}%{c|cccc|cccc|cccc|cccc|c}
\hline
motion & \multicolumn{4}{c}{Waiting} & \multicolumn{4}{c}{Walking Dog}& \multicolumn{4}{c}{Walking Together} & \multicolumn{4}{c|}{Average}&\multirow{2}{*}{\#Params(M)}\\
\cline{1-17}%\hline
time(ms)&80&160&320&400&80&160&320&400&80&160&320&400&80&160&320&400 \\
\hline
ResSup\cite{03} & 29.5& 60.5& 119.9& 140.6& 60.5& 101.9& 160.8& 188.3& 23.5& 45.0& 71.3& 82.8& 30.8& 57.0& 99.8& 115.5&\qquad--\\
ConvS2S\cite{04} &17.9& 36.5& 74.9& 90.7& 40.6& 74.7& 116.6& 138.7& 15.0& 29.9& 54.3& 65.8& 19.6& 37.8& 68.1& 80.2&\qquad--\\
%\hline
LTD\cite{05}& 9.5& 22.0& 57.5& 73.9& 32.2& 58.0& 102.2& 122.7 & 8.9& { 18.4}& 35.3& 44.3& 12.1& 25.0& 51.0& 61.3&\quad$\approx$ 2.55\\
\hline
Ours&{\bf 7.6}&{\bf 19.7}&{\bf 53.6}&{\bf 69.3} & {\bf 23.5}&{\bf 51.9}&{\bf 98.4}&{\bf 116.7}&  {\bf 7.5}&{\bf 16.5}&{\bf 32.2}&{\bf 42.2}&{\bf 10.1}&{\bf 22.7}&{\bf 47.7}&{\bf 58.6}&$\quad\approx$ \bf 0.28\\
\hline
\end{tabular}
\end{center}
\label{results_h36mshort}
\vspace{-2.5em}
\end{table*}

\begin{table*}[!t] % long-term results on h36m
\vspace{0.5em}
\caption{Long-term prediction on H$3.6$M.}
\scriptsize
\begin{center}
\begin{tabular}{c|cc|cc|cc|cc|cc|cc|cc|cc}%{p{.9cm}p{0.18cm}p{0.18cm}p{0.3cm}p{0.3cm}p{0.3cm}p{0.35cm}|p{0.18cm}p{0.18cm}p{0.3cm}p{0.3cm}p{0.3cm}p{0.35cm}|p{0.18cm}p{0.18cm}p{0.3cm}p{0.3cm}p{0.3cm}p{0.35cm}|p{0.18cm}p{0.18cm}p{0.3cm}p{0.3cm}p{0.3cm}p{0.35cm}}
\hline
motion & \multicolumn{2}{c}{Walking} & \multicolumn{2}{c}{Eating}& \multicolumn{2}{c}{Smoking} & \multicolumn{2}{c}{Discussion}
&\multicolumn{2}{c}{Directions} & \multicolumn{2}{c}{Greeting}& \multicolumn{2}{c}{Phoning} & \multicolumn{2}{c}{Posing}\\
\hline
time(ms)&560 &1000&560 &1000&560 &1000&560 &1000&560 &1000&560 &1000&560 &1000&560 &1000\\
\hline
LTD\cite{05}&42.2&51.3&56.5&{\bf 68.6}&{\bf 32.3}&{\bf 60.5}&{\bf 70.4}&103.5&85.8&109.3&{\bf 91.8}&{\bf 87.4}&65.0&113.6&113.4&220.6\\
\hline
Ours&{\bf 40.9}&{\bf 42.8}&{\bf 56.2}&70.2&34.7	&61.1&	78.5	&{\bf 103.3}&{\bf 	82.4}&{\bf 102.8}&96.3	&92.4&{\bf 62.4}&	{\bf 109.3}&{\bf 111.3}&{\bf 207.6}\\
\hline
\end{tabular}
\begin{tabular}{p{1cm}|p{0.3cm}p{0.45cm}|p{0.25cm}p{0.45cm}|p{0.35cm}p{0.45cm}|p{0.5cm}p{0.45cm}|p{0.4cm}p{0.45cm}|p{0.4cm}p{0.5cm}|p{0.85cm}p{0.45cm}|p{0.4cm}p{0.45cm}|p{1cm}}%{c|cc|cc|cc|cc|cc|cc|cc|cc|c}
\hline
motion & \multicolumn{2}{c}{Purchases} & \multicolumn{2}{c}{Sitting}& \multicolumn{2}{c}{Sitting down} & \multicolumn{2}{c}{Taking photo}
&\multicolumn{2}{c}{Waiting} & \multicolumn{2}{c}{Walking Dog}& \multicolumn{2}{c}{Walking Together} &\multicolumn{2}{c|}{Average}&\multirow{2}{*}{\#Params(M)}\\
\cline{1-17}
time(ms)&560 &1000&560 &1000&560 &1000&560 &1000&560 &1000&560 &1000&560 &1000&560 &1000\\
\hline
LTD\cite{05}&94.3&{\bf 130.4}&79.6&114.9&{\bf 82.6}&140.1&{\bf 68.9}&{\bf 87.1}&100.9&167.6&136.6	&{\bf 174.3}&57.0&85.0&78.5&114.3&\quad$\approx$ 2.55\\
\hline
Ours&{\bf 90.9}&130.6&{\bf 74.7}&{\bf 112.3}&86.1&{\bf 134.4}&74.4&97.9&{\bf 94.2}&{\bf 162.2}&{\bf 133.8}&174.5&{\bf 55.7}&{\bf 76.7}&{\bf 78.2}&{\bf 111.9}&\quad$\approx$ {\bf 0.28}\\
\hline
\end{tabular}
\end{center}
\label{results_h36mlong}
\vspace{-2.0em}
\end{table*}

\subsection{Datasets and implementation details}

\textbf{Datasets. }(1) Human $3.6$M (H$3.6$M) \cite{16} is the most commonly used dataset in the field of human motion prediction. For a fair comparison, we adopt the exactly same data preprocessing methods as in \cite{03,05}, such as sampling frequency, training/test sets splits, etc. After processing, each skeletal frame is with 22 joints, and each joint represented by a 3D coordinate, and then one frame is represented as a 66-dimensional vector, \ie $N$=66 for H$3.6$M dataset. (2) For CMU mocap dataset (CMU-Mocap), the counterpart $N$ is 75. Consistent with the baselines \cite{05}, the data preprocessing on CMU-Mocap is carried out on our experiments for fair comparison.

\textbf{Implementation details. }All experiments are conducted by using PyTorch framework \cite{17} on one GTX1080TI GPU. ADAM \cite{18} optimizer is used to train our model. The batch size is 32. The learning rate is initialized to 0.001 decayed with factor of 0.98 for every 1 epochs. And the gradient was clipped to a maximum L2-norm of 1. To train our model, Mean Per Joint Position Error (MPJPE) \cite{16} loss with bone length constraint \cite{19} is employed:
\begin{equation}
 \begin{split}
loss = \frac{1}{JT'}{\sum\limits_t^{T'} {\sum\limits_j^J {\left\Vert{{\mathbf{J}}_{tj}} - {{\mathbf{{\hat J}}_{tj}}} \right\Vert}_2 }} \\+ \lambda \frac{1}{BT'}{\sum\limits_t^{T'} {\sum\limits_b^B {\left| {{l_b} - {{\hat l}_{tb}}} \right|}}}
 \end{split}
\end{equation}
where ${{\mathbf{{\hat J}}}_{tj}}$ is a 3-dimensional vector and denotes the coordinate of $j$th joint in predicted frame $t$, ${{\mathbf{J}}_{tj}}$ the ground-truth counterpart. And the second part of the above loss is bone length constraint, in which ${{\hat l}_{tb}}$ is a number and denotes the length of $b$th bone in predicted frame $t$, and $l_b$ the ground-truth length of the $b$th bone. In our experiments, $\lambda$ is set to 0.1.

%\subsubsection{}
%\subsection{Baselines}

%We compare the proposed model with three recent works in the 3D Cartesian system, including recurrent model ResSup. \cite{03}, CNN-based feedforward model ConvS2S \cite{04}, and GCN-based feedforward model LTD \cite{05}. Note that all results of the baselines can be found in \cite{05}.

\subsection{Comparison with the state-of-the-art}

\begin{table*}[!t] % results on cmu
\caption{Short and long-term prediction on CMU-mocap.}
\scriptsize
\begin{center}
\begin{tabular}{c|ccccc|ccccc|ccccc}%{p{.9cm}p{0.18cm}p{0.18cm}p{0.3cm}p{0.3cm}p{0.35cm}|p{0.18cm}p{0.18cm}p{0.3cm}p{0.3cm}p{0.35cm}|p{0.18cm}p{0.18cm}p{0.3cm}p{0.3cm}p{0.35cm}
\hline%%%%%%%%%%%
motion& \multicolumn{5}{c}{Basketball} & \multicolumn{5}{c}{Basketball Signal}& \multicolumn{5}{c}{Directing Traffic}\\
\hline
time (ms) & 80 &160 & 320 &400 &1000& 80 &160 & 320 &400 &1000& 80 &160 & 320 &400 &1000\\
\hline
LTD\cite{05}&14.0&25.4 &	49.6 &61.4&106.1	&3.5 &	6.1 &	11.7 &{\bf 15.2}&{\bf 53.9}&	7.4 &	15.1 &31.7 &	42.2 &152.4\\
\hline
Ours&{\bf 10.6}&{\bf 19.6}&{\bf 42.4}&{\bf 53.9}&{\bf 99.0}&{\bf 2.5}&{\bf 3.9}&{\bf 11.0}&{\bf 15.2}&68.6&{\bf 5.8}&	{\bf 10.4}&{\bf 23.5}&{\bf 31.4}&{\bf 132.2}\\
\hline%%%%%%%%%%%%%%%%
\hline%%%%%%%%%%%%%%%%
motion& \multicolumn{5}{c}{Jumping} & \multicolumn{5}{c}{Running}& \multicolumn{5}{c}{Soccer}\\
\hline
time (ms) & 80 &160 & 320 &400 &1000& 80 &160 & 320 &400 &1000& 80 &160 & 320 &400 &1000\\
\hline
LTD\cite{05}&16.9& 	34.4 &	76.3 	&{\bf 96.8}&164.6	&25.5 	&36.7& 	39.3 &	39.9 	&{\bf 58.2}&11.3 	&21.5 	&44.2& 	55.8 &	117.5\\
\hline
Ours&{\bf 11.6}&{\bf 26.9}&{\bf 73.2}&	99.0&{\bf 152.5}	&{\bf 18.3}&{\bf 23.5}&{\bf 26.8}&{\bf 33.7}&	73.4	&{\bf 7.4	}&{\bf 16.1}&{\bf 39.0	}&{\bf 51.5}&{\bf 108.7}\\
\hline
\end{tabular}
\begin{tabular}{c|ccccc|ccccc|ccccc|c}%{p{.9cm}p{0.18cm}p{0.18cm}p{0.3cm}p{0.3cm}p{0.35cm}|%{c|ccccc|ccccc|ccccc|c}
\hline%%%%%%%%%%%
motion& \multicolumn{5}{c}{Walking} & \multicolumn{5}{c}{Wash Window}& \multicolumn{5}{c|}{Average}&\multirow{2}{*}{\#Params(M)}\\
 \cline{1-16}
time (ms) & 80 &160 & 320 &400 &1000& 80 &160 & 320 &400 &1000& 80 &160 & 320 &400 &1000\\
\hline
LTD\cite{05}&7.7 &	11.8 &	19.4 &	23.1 &{\bf 40.2}&5.9 &	11.9 &	30.3 &	40.0 &{\bf 79.3}&11.5 &	20.4 &	37.8 &	46.8 &	96.5&$\approx$ 2.70\\
\hline
Ours&{\bf 6.8	}&{\bf 10.4}&{\bf 17.8}&{\bf 20.2}&	43.8&{\bf 4.8}	&{\bf 9.3	}&{\bf 27.8}&{\bf 37.4}&79.7	&{\bf 8.5}&{\bf 15.0}&{\bf 32.7}&{\bf 42.8}	&{\bf 94.7}&$\approx$ {\bf 0.30}\\
\hline
\end{tabular}
\end{center}
\label{results_cmu}
\vspace{-2.5em}
\end{table*}

\subsubsection{Comparison on H$3.6$M}

\textbf{Short-term motion prediction. }We first report our results on short-term prediction. Note that short-term prediction is to predict the poses within the next 400 milliseconds, \ie next 10 frames. Table \ref{results_h36mshort} reports the MPJPE (in millimeters) of short-term prediction for all 15 activities and the average MPJPE on the 15 activities. It can be observed that our method outperforms all the baselines on the average MPJPE, which demonstrates the effectiveness of our method. From the view of each action, although there is a narrow gap between our method and LTD \cite{05} for some actions (i.e. ``Walking'', ``Eating'', ``Discussio'', ``Phoning'' and ``Taking photo''), our results are still competitive in most of the actions. Furthermore, for actions with complex movements (like ``Walking Dog'' and ``Greeting''), our model surpasses the baselines by a large margin (\eg compared with the start-of-the-art, the average error of our predictions decreasing 6mm for ``Walking Dog'' at 400ms, and for ``Greeting'' the error decreasing 8.1mm).

\begin{figure}[!h]  % frame-wise visualization results
\begin{center}
\subfigure[Walking dog]{\includegraphics[width=0.5\textwidth]{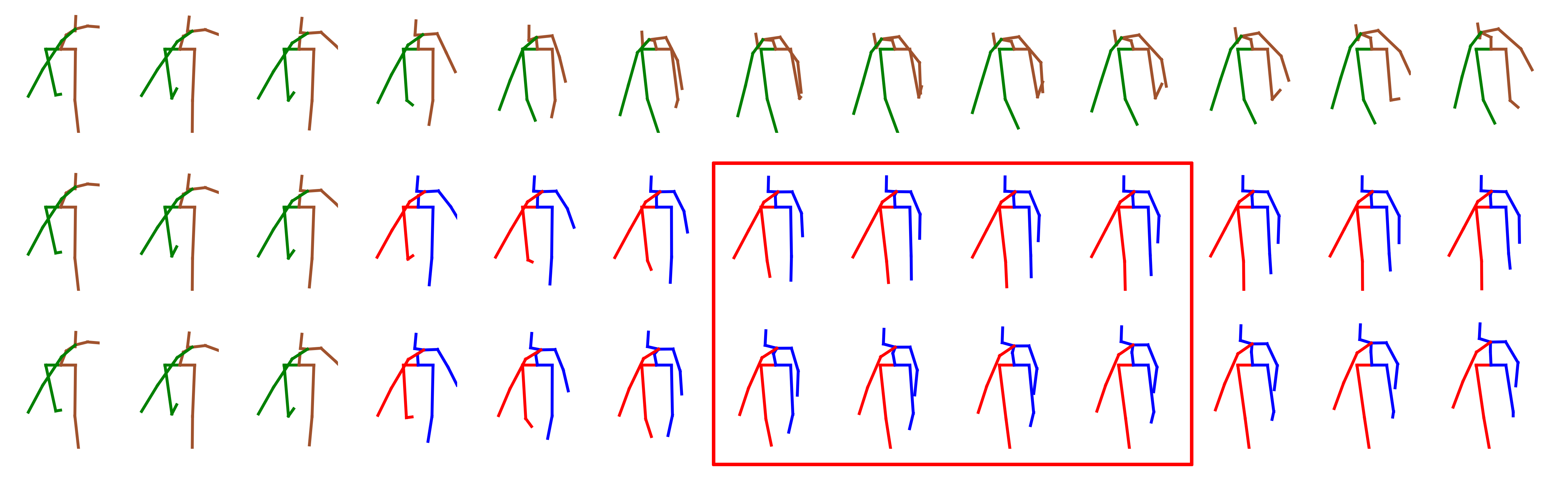}%[width=0.5\textwidth, trim = 100mm 1mm 0mm 0mm, clip=true]{myfig/vis1.eps} %[width=0.98\columnwidth,height=0.6in,trim = 52mm 18mm 48mm 18mm, clip=true]{vis1.eps}
\label{h36m_walking}
}\vspace{-0.5em}
\subfigure[Purchases]{\includegraphics[width=0.5\textwidth]{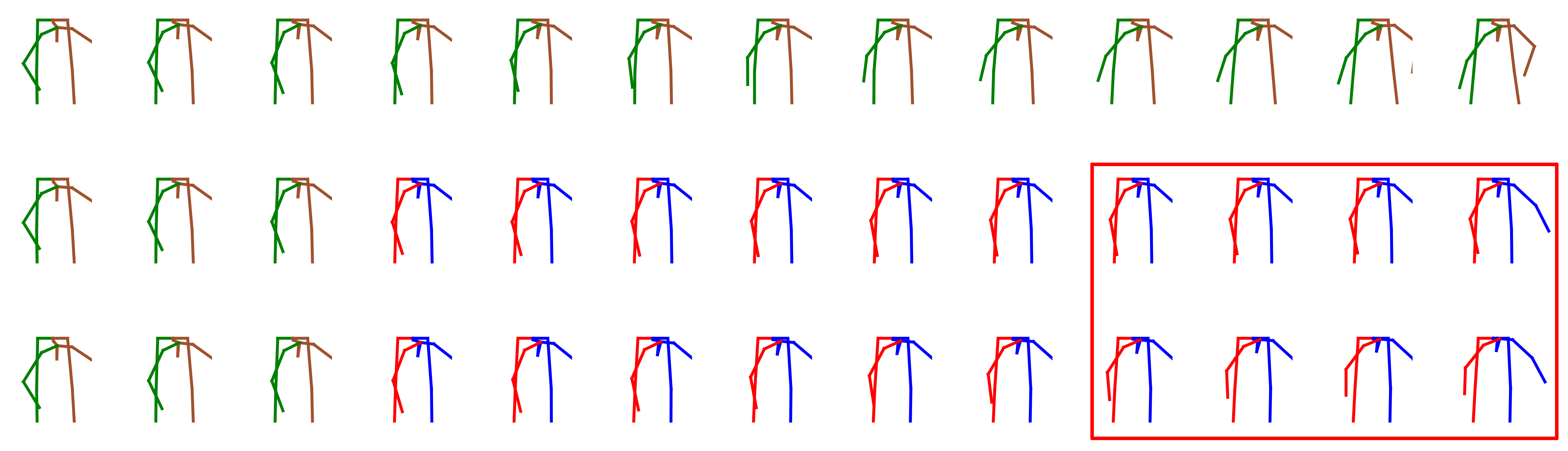}%[width=0.5\textwidth, trim = 100mm 1mm 0mm 0mm, clip=true]{myfig/vis2.eps}%[width=0.98\columnwidth,height=0.6in,trim = 52mm 18mm 48mm 29mm, clip=true]{vis1.eps}
\label{h36m_pur}
}%\vspace{-0.5em}

\subfigure[Waiting]{\includegraphics[width=0.5\textwidth]{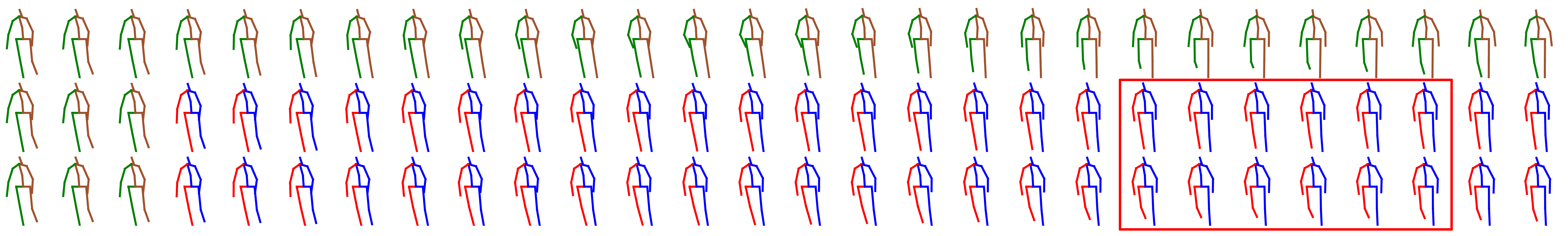}%[width=0.5\textwidth, trim = 100mm 1mm 0mm 0mm, clip=true]{myfig/vis2.eps}%[width=0.98\columnwidth,height=0.6in,trim = 52mm 18mm 48mm 29mm, clip=true]{vis1.eps}
\label{h36m_wait}
}
\end{center}
\caption{Qualitative comparison of short-term predictions (``Walking dog'' and ``Purchases'') and long-term predictions (``Waiting'') on H$3.6$M. In each sub-figure, from top to bottom, the ground-truth, the predictions of LTD \cite{05} and our predictions is shown. The poses in red-blue represent the predicted ones, others are ground-truth. Our predictions are with higher dynamics and the evolutions are more near to the ground-truth.}
\label{h36m_vis}
\vspace{-1em}
\end{figure}

\textbf{Long-term motion prediction. }Table \ref{results_h36mlong} illustrates the long-term results of our model and the state-of-the-art method LTD \cite{05}. Our model achieves slightly better performance on average and most of the motions with much lighter weight. Indeed, the advantage is not as obvious as in short-term prediction; and the reason is mainly that the long-term prediction needs higher model capacity to model the spatiotemporal dependencies than the short-term prediction. Considering that the number of our network parameters is only about 0.11 times of LTD, the proposed method is still competitive in long-term prediction.

\textbf{Comparison of model size. }More importantly, our model has great advantages in the comparison of the number of parameters. As shown in Table \ref{results_h36mshort}, the parameters of  LTD \cite{05}  are about 2.55 million, and the parameters of our model is about 0.28 million. The proposed model achieves the lowest MPJPE on average and most of cases with a much smaller model size (with less than 0.11 times the number of parameters), which fully shows the strong competitiveness of our method and proves the efficiency and effectiveness of our model for spatiotemporal modeling. Namely, our model has found a balance between the prediction quality and the complexity of network, which shows the a promising application prospect for the devices with limited computational resources.

\textbf{Qualitative comparison. }We also make qualitative comparison with the start-of-the-art LTD \cite{05} by visualizing the predicted motion on H$3.6$M. As can be seen in Figure \ref{h36m_vis}, compared with the baseline, our predictions are with higher dynamics and more near to the ground-truth. In detail, as shown in Figure \ref{h36m_walking}, in the red boxes, the leg in blue changes slowly and tends to converge to mean pose in the predicted results of LTD, while the evolutions of our predictions are closer to the ground-truth with higher dynamics. Similar experimental phenomenon can be observed in Figure \ref{h36m_pur} and Figure \ref{h36m_wait}. It shows that our model can capture the dynamic information more efficiently, which further reflects the effectiveness of our model.

\subsubsection{Comparison on CMU mocap. }

Table \ref{results_cmu} reports the results on CMU mocap dataset. The results contain two parts, short-term prediction (to 400ms) and long-term prediction (in 1000ms). Note that the mirror transformation and bone length loss is not used because of the prediction on CMU mocap is not difficult like that on H$3.6$M. And even so, as is reported in Table\ref{results_cmu}, our model outperforms the state-of-the-art on average MPJPE and most of activities with less parameters. This further prove the effectiveness and efficiency of our model.

\subsection{Ablation analysis}

In this section, we conduct comprehensive ablation experiments on H$3.6$M dataset to evaluate some crucial components of our method, including sub-joint trajectory stream (SJTS), joint trajectory stream (JTS) and mirror transformation (MT). Table \ref{aba} reports the average MPJPE of short-term prediction and the size of networks.
\vspace{-0.5em}
\begin{table}[h]
\caption{Ablation results on H$3.6$M.}
\vspace{0.3em}
\footnotesize
\label{table_example}
\begin{center}
\begin{tabular}{c|cccc|c}
\hline
Time(ms)&80&160&320&400&\#Parmas(M)\\
\hline
w/o SJTS (drop)&	10.6	&24.4	&51.2	&61.9	&$\approx$ 0.13\\
%\hline
w/o SJTS (rep)&	10.4	&23.6	&49.5	&60.2	&$\approx$ 0.23\\
%\hline
w/o JTS (drop)&	10.1	&23.1	&48.8	&59.8	&$\approx$ 0.18\\
%\hline
w/o JTS (rep)&	10.2	&23.1	&48.9	&59.1	&$\approx$ 0.33\\
%\hline
w/o MT &	10.3&	23.4	&49.6	&60.4	&$\approx$ 0.28\\%10.3&	23.6	&50.4	&61.7	&$\approx$ 0.28\\
%\hline10.3 23.4 49.6 60.4
Proposed	&10.1	&22.7	&47.7	&58.6	&$\approx$ 0.28\\
\hline
\end{tabular}
\end{center}
\vspace{-1.5em}
\label{aba}
\end{table}

\textbf{The evaluation of SJTS. }In order to verify the SJTS, we design two ablation experiments: (1) directly removing the SJTS from the proposed model, marked as ``w/o SJTS (drop)''; (2) replacing the SJTS with the JTS in the proposed model, marked as ``w/o SJTS (rep)''. From the comparison of the ``Proposed'' and ``w/o SJTS (drop)", the errors increase when removing the SJTS, which indicates that the SJTS is helpful to improve performance. To reduce the influence of the decreasing of parameters caused by the direct removal of the SJTS, we replace the SJTS with the JTS in the proposed model. And from the comparison of the ``Proposed'' and `` w/o SJTS (rep)'', the errors also grow. In summary, the prediction errors get larger without the SJTS. One reason is that the model without the SJTS cannot flexibly capture spatiotemporal correlations between global sub-joint trajectories, which shows that the fine-grained spatiotemporal features are particularly crucial for human motion prediction and also reflects the effectiveness of the SJTS.

\textbf{The evaluation of JTS. }To validate the effectiveness of the JTS, similar to the above ablation experiments, we conduct two experiments: (1) we directly remove the JTS, denoted as `` w/o JTS (drop)''; (2) replacing the JTS with the SJTS, denoted as `` w/o JTS (rep)''. From the experimental results of the `` w/o JTS (drop)'', `` w/o JTS (rep)'' and ``Proposed'', similar conclusions can be drawn, \ie the prediction quality declines without the JTS. One explanation could be that the spatiotemporal modeling within the JTS is subject to the constraints of the skeletal prior construction, which makes full use of the structural characteristics of the skeleton to carry out hierarchical spatiotemporal modeling.

%In summary, the SJTS and the JTS are both beneficial to improve the prediction performance. The former can flexibly explore spatiotemporal correlations with no spatial constraints, while the latter is with the ability of hierarchical spatiotemporal modeling under the constraints of prior construction. The non-constrained and constrained spatiotemporal information can be complementary by the fusion of the two streams, which is crucial to improve the prediction performance.

\textbf{The evaluation of MT. }To study the effectiveness of the MT, we remove the operation and only use the original skeleton data for training, the experiment marked as `` w/o MT''. As shown in Table \ref{aba}, compared with the "Proposed'' and ``w/o MT'', it can be seen that the prediction error is significantly increased after the MT is discarded, which indicates the effectiveness of the strategy. The main reasons may be that, the model tends to learn biases about the habit-related motion patterns without the MT, such as right-handedness, causing with the poor generalization ability to new subjects.

\section{CONCLUSIONS}

In this paper, we have represented human motion as multi-grained trajectories and proposed a multi-grained trajectory GCN based, effective and efficient method for habit-unrelated human motion prediction. Unlike the previous works which aim for low error of the prediction results, our method not only pays attention to the high accuracy, but also the high efficiency and the lightweight of the prediction model. In order to obtain the accuracy and the efficiency, on the one hand, the proposed method mines the spatial-temporal correlations based on the coarse-grained joint trajectories and fine-grained sub-joint trajectories. On the other hand, a new motion generation scheme is proposed to generate the motion with left-handedness, to better model the motion with less bias to the human habit. Thanks to the motion generation scheme and the proposed GCN framework, our model achieves state-of-the-art performance with competitive model size.

%-------------------------------------------------------------------------

{\small
\bibliographystyle{ieee_fullname}
\bibliography{egbib}
}

\end{document}